\documentclass[11pt,a4paper]{article}
\usepackage[hyperref]{acl2019}
\usepackage{times}
\usepackage{latexsym}
\usepackage{graphicx}
\usepackage{url}
\usepackage{caption}
\usepackage{amsmath}
\usepackage{subcaption}
\usepackage{algorithmic}
\usepackage{dcolumn}
\newcommand{\kh}[1]{\textcolor{red}{}}
\newcommand{\al}[1]{\textcolor{blue}{}}

\newcommand{\khdone}[1]{\textcolor{green}{}}

\newcolumntype{L}{D{.}{.}{2,4}}

\aclfinalcopy % Uncomment this line for the final submission
\title{Making Asynchronous Stochastic Gradient Descent Work for Transformers}

\author{Alham Fikri Aji \and
  Kenneth Heafield \\
  School of Informatics, University of Edinburgh \\
  10 Crichton Street \\
  Edinburgh EH8 9AB \\
  Scotland, European Union \\
  \texttt{a.fikri@ed.ac.uk, kheafiel@inf.ed.ac.uk}
}

\date{}

\begin{document}
\maketitle
\begin{abstract}
Asynchronous stochastic gradient descent (SGD) is attractive from a speed perspective because workers do not wait for synchronization.  However, the Transformer model converges poorly with asynchronous SGD, resulting in substantially lower quality compared to synchronous SGD.  To investigate why this is the case, we isolate differences between asynchronous and synchronous methods to investigate batch size and staleness effects. We find that summing several asynchronous updates, rather than applying them immediately, restores convergence behavior. With this hybrid method, Transformer training for neural machine translation task reaches a near-convergence level 1.36x faster in single-node multi-GPU training with no impact on model quality.
\end{abstract}

\section{Introduction}

%Start with transformer model.
% Prior work (Scaling Neural Machine Translation and Best of Both Worlds) reports that synchronous SDG has higher quality than async
%In this paper, we investigate why

Models based on Transformers~\cite{vaswani2017attention} achieve state-of-the-art results on various machine translation tasks~\cite{wmt2018finding}.  Distributed training is crucial to training these models in a reasonable amount of time, with the dominant paradigms being asynchronous or synchronous stochastic gradient descent (SGD).  Prior work \cite{chen2016revisiting, chen2018best, ott2018scaling} found that asynchronous SGD yields low quality models, but did not elaborate further.  We confirm this experimentally in Section~\ref{motivational}.  Then we conduct ablation studies to understand what makes asynchronous SGD under-perform, leading to a hybrid that trains high-quality models without waiting for synchronization barriers.  

% Distributed stochastic gradient descent (SGD) is often used to speed up the training process of neural machine translation (NMT)~\cite{bahdanau2014neural}. An asynchronous variant of distributed SGD can be chosen for additional speed boost by potentially sacrificing quality~\cite{zhang2015fast, chen2016revisiting}. However, we found that damage to the quality is more severe in the Transformer~\cite{vaswani2017attention} model. Our finding is consistent with other reports~\cite{chen2016revisiting, ott2018scaling}.

In synchronous SGD, gradients are collected from all workers and summed before updating, equivalent to one large batch. These accumulation and waiting processes are absent in asynchronous SGD, where updates are applied immediately after they are computed by any processor.  Since each update comes from one processor, the batch size per update in asynchronous SGD is smaller.  Prior work has shown that smaller batches degrade the final quality of Transformers \cite{smith2017don, popel2018training}. Moreover, the model has typically updated several times while a gradient was computed, so gradients are stale.  Stale gradients potentially degrade final model quality \cite{zhang2015staleness, srinivasan2018analysis}.  

We investigate the effect of batch size and stale gradients on Transformer training, comparing with recurrent neural network (RNN) training.  All of these experiments use the Adam optimizer, which has shown to perform well on a variety of tasks \cite{kingma2014adam} and was used in the original Transformer paper \cite{vaswani2017attention}.  We find that small batch sizes slightly degrade quality while stale gradients substantially degrade quality.  

We adopt prior work that summed gradients in various contexts \cite{dean2012large, lian2015asynchronous, ott2018scaling, bogoychev2018accelerating} to increase the batch size while reducing staleness.  Empirically, summing gradients globally in the parameter server performs equally well with synchronous SGD in terms of BLEU score, while also maintaining the speed benefit of asynchronous SGD.

% Our contributions in this paper are as follows:
% \begin{itemize}
%    \item We empirically confirm that asynchronous SGD severely damages Transformer model, compared to an RNN. \kh{Really, our first contribution is to ``confirm'' something?}
%    \item We perform ablation studies, showing that both batch size and staleness were affecting the Transformer's learning capability, in which the latter is more prominent.
%    \item We test several hybrids of asynchronous and synchronous SGD, finding a combination with quality and efficiency.  
%\end{itemize}

\section{Exploring Asynchronous SGD}
In this section, we analyze the causes of poor performance in asynchronous SGD, including experiments.

\subsection{Baseline: The Problem}
\label{motivational}
To motivate this paper and set baselines, we first measure how poorly Transformers perform when trained with baseline asynchronous SGD~\cite{chen2016revisiting, chen2018best, ott2018scaling}.  We train a Transformer model under both synchronous and asynchronous SGD, contrasting the results with an RNN model.  Moreover, we sweep learning rates to verify this effect is not an artifact of choosing hyperparameters that favor one scenario.

\begin{figure*}[t!]
\centering
\begin{subfigure}[b]{.45\textwidth}
\centering
\includegraphics[height=4.2cm]{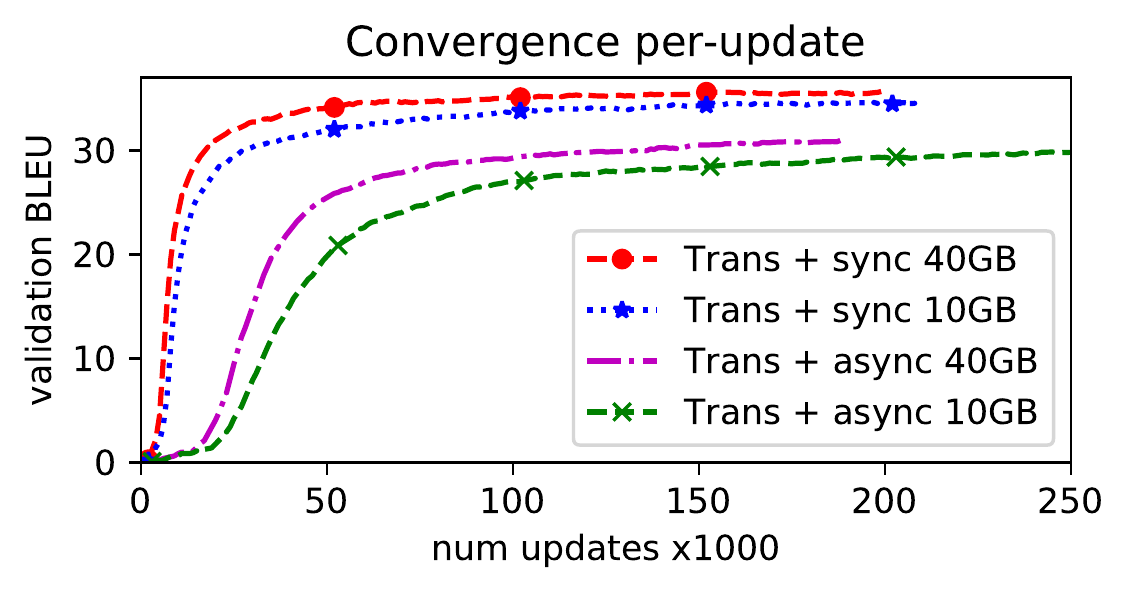}
\includegraphics[height=4.2cm]{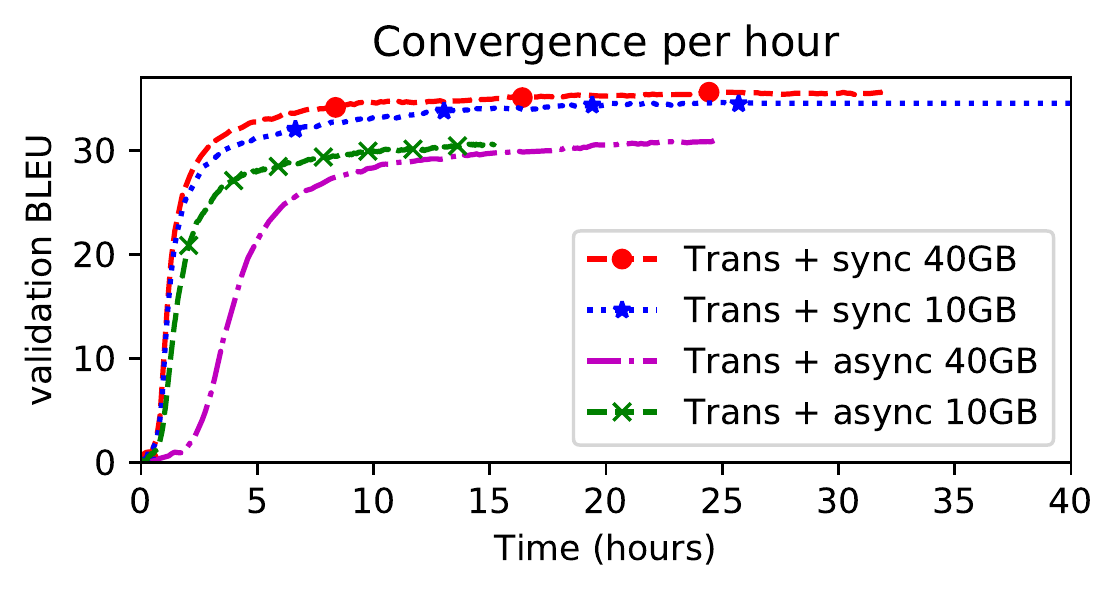}
\caption{Convergence over updates and time in Transformer model with various batch sizes}
\label{fig:trans_vs_batch}
\end{subfigure}
\qquad
\begin{subfigure}[b]{.45\textwidth}
\centering
\includegraphics[height=4.2cm]{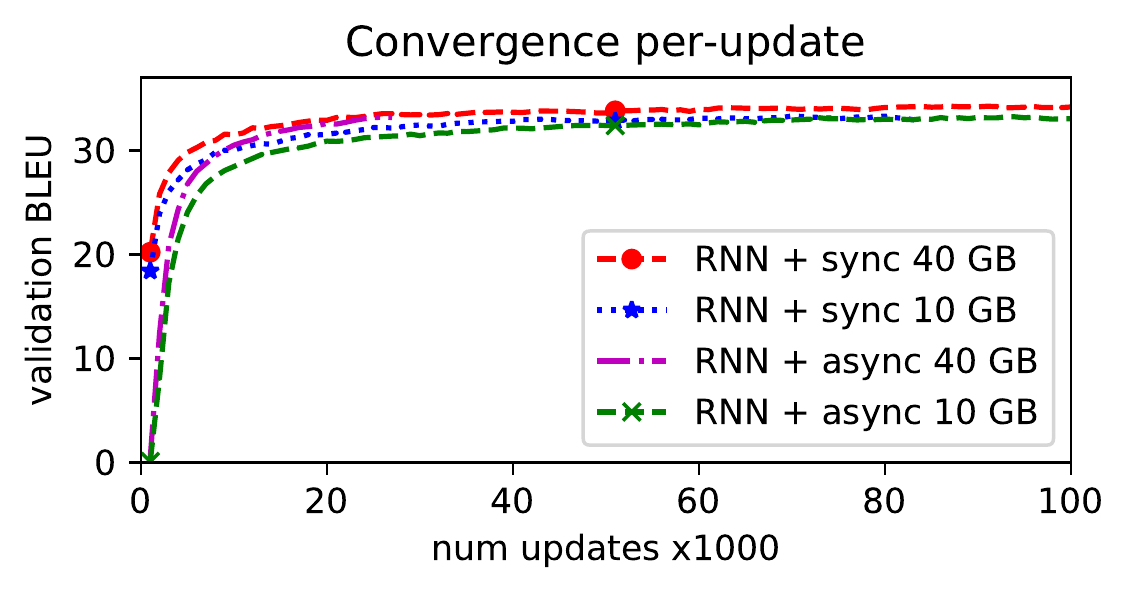}
\includegraphics[height=4.2cm]{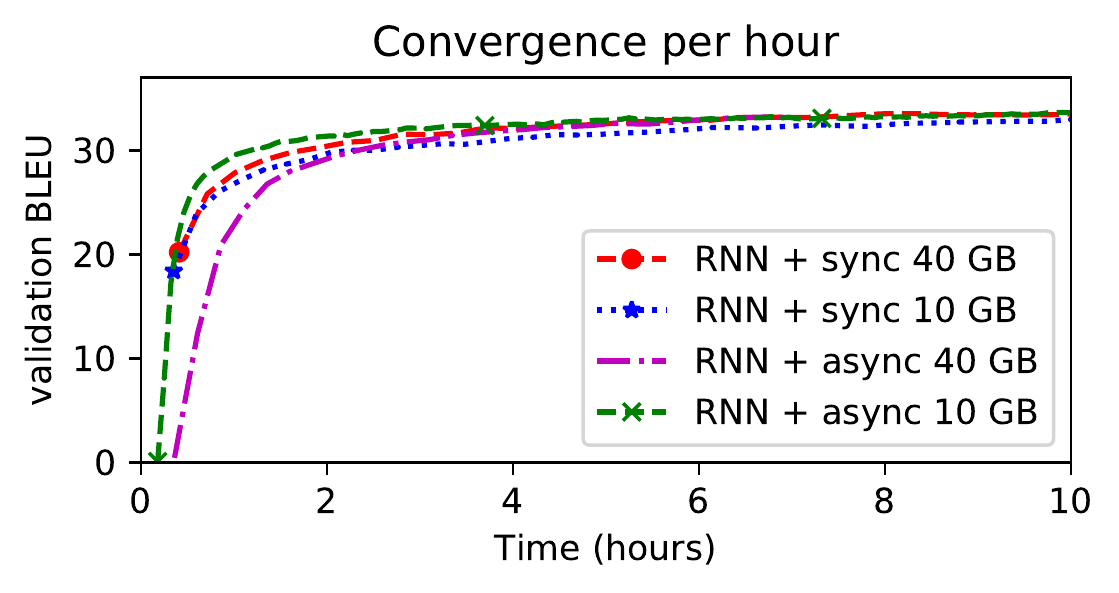}
\caption{Convergence over updates and time in RNN model with various batch sizes}
\label{fig:rnn_vs_batch}
\end{subfigure}
\caption{The effect of batch sizes on convergence of Transformer and RNN models.
}
\label{fig:all_vs_batch}
\end{figure*}

Our experiments use systems for the WMT 2017 English to German news translation task. The Transformer is standard with six encoder and six decoder layers. The RNN model~\cite{barone2017deep} is based on the University of Edinburgh's winning WMT17 submission \cite{sennrich2017university} and has 8 layers. Both models use back-translated monolingual corpora~\cite{sennrich2015improving} and byte-pair encoding~\cite{sennrich2015neural}. Model performance is validated on newstest2016; we preserve newstest2017 as test for later experiments.

\begin{table}[h!]
\begin{tabular}{lrrrr}
\hline
  & \multicolumn{2}{c}{Trans. BLEU} & \multicolumn{2}{c}{RNN BLEU}  \\ 
 Learn Rate & Sync. & Async. & Sync. & Async. \\ \hline
 0.0002 & 35.08 & 13.27 & 34.11 & 33.77 \\
 0.0003 & 35.66 & 30.72 & 33.79 & 33.95 \\ 
 0.00045 & 35.59 & 5.21 & 33.68 & 33.68 \\ 
 0.0006 & 35.42 & 0.00 & 34.30 & 33.76 \\ 
 0.0009 & 34.79 & 0.00 & 34.28 & 33.47 \\ 
 0.0012 & 33.96 & 0.00 & 34.37 & 33.23 \\
 0.0024 & 29.35 & 0.00 & 33.98 & 32.83 \\
 0.00375 & 25.25 & 0.00 & 33.80 & 31.89 \\
 \hline
\end{tabular}
\caption{Performance of the Transformer and RNN model trained synchronously and asynchronously, across different learning rates.}
\label{table:lr-sweep}
\end{table}

We follow the rest of the hyperparameter settings on both Transformer and RNN models as suggested in the papers \cite{vaswani2017attention,sennrich2017university}. We trained our model in a four GPUs environment with a dynamic batch size of 10 GB for each GPU with the Marian toolkit \cite{junczys2018marian}. The models are trained for 8 epochs or until reaching five continuous validations without loss improvement. Quality is measured on newstest2016 using sacreBLEU~\cite{post2018call}, preserving newstest2017 as test for later experiments. The Transformer's learning rate is linearly warmed up for 16k updates. We apply an inverse square root learning rate decay following~\newcite{vaswani2017attention} for both models. Parameters are optimized using Adam~\cite{kingma2014adam} with $\beta_1 = 0.9$ and $\beta_2 = 0.98$.

Results in Table~\ref{table:lr-sweep} confirm that asynchronous SGD generally yields lower-quality systems than synchronous SGD. For Transformers, the asynchronous results are catastrophic, often yielding 0 BLEU. We can also see that Transformers and asynchronous SGD are more sensitive to learning rates compared to RNNs and synchronous SGD.

For subsequent experiments, we will use a learning rate of 0.0003 for Transformers and 0.0006 for RNNs.  These were near the top in both asynchronous and synchronous settings (Table~\ref{table:lr-sweep}).

\kh{I commented out the background section.  Felt it wasn't adding anything.}

\subsection{Batch Size}

Synchronous SGD has a larger effective batch size because it sums gradients from all workers, hence approximates the gradient better. This section investigates the extent to which batch size is the cause of poor convergence.  

We use dynamic batching, where we fit as many sentences as it can into a fixed amount of memory (so e.g.\ more sentences will be in a batch if all of them are short), hence batch sizes are denominated in memory sizes. Our GPUs each have 10 GB available for batches which, on average, corresponds to 250 sentences.

Since there are 4 GPUs, baseline synchronous SGD has an effective batch size of 40 GB, compared to 10 GB in asynchronous.  We fill in the two missing scenarios: synchronous SGD with a total effective batch size of 10 GB and asynchronous SGD with a batch size of 40 GB.  Because GPU memory is limited, we simulate a larger batch size in asynchronous SGD by locally accumulating gradients in each processor four times before sending the summed gradient to the parameter server \cite{ott2018scaling, bogoychev2018accelerating}.

Using a larger batch size reduces noise in estimating the overall gradient~\cite{wang2013variance}. Therefore, both models achieved slightly better BLEU \emph{per update} in their early stage of training as shown in Figure~\ref{fig:all_vs_batch}.  However, serially computing batches is time consuming so asynchronous training with a 40 GB batch size performs worse in terms of BLEU by time.

\newcite{goyal2017accurate} suggested that the learning rate can be increased proportionate to the batch size to cope with slower processing time.  We can scale up the learning rate in RNN training from 0.0006 to 0.0012 without reducing the final quality. Unfortunately, increasing the learning rate further causes quality degeneration in the Transformer model.

From this experiment, we conclude that batch size is not the primary driver of poor performance of asynchronously trained Transformers, though it does have some lingering impact on final model quality.  For RNNs, batch size and distributed training algorithm had little impact beyond the early stages of training, continuing the theme that Transformers are more sensitive to noisy gradients.  

\subsection{Gradient Staleness}
\label{staleexperiment}

\begin{figure*}[t!]
\centering
  \begin{subfigure}[b]{0.45\textwidth}
    \centering
    \includegraphics[height=4.2cm]{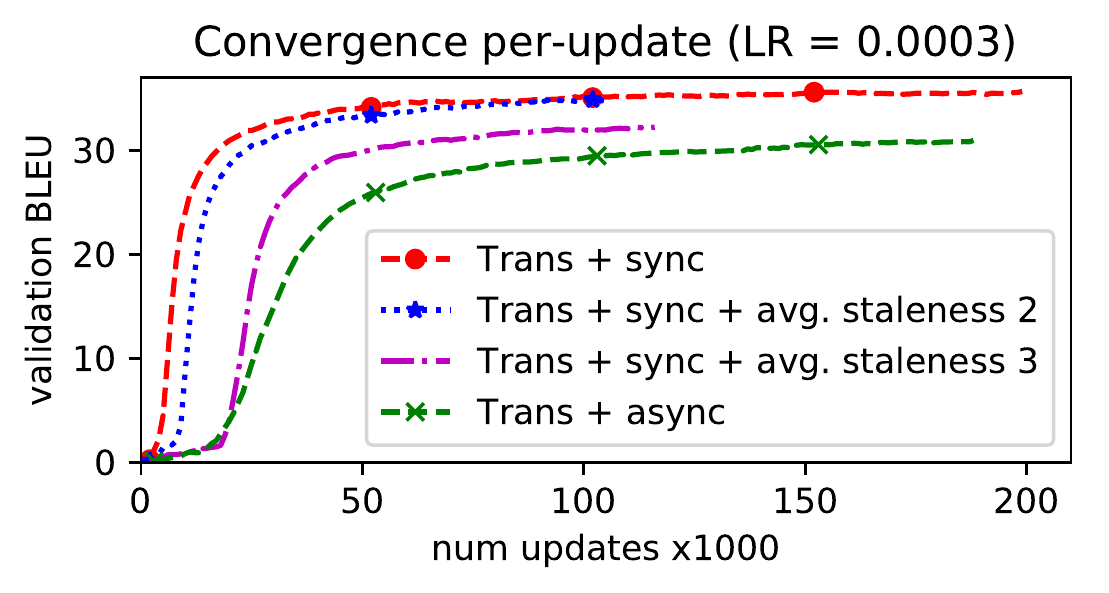}
    \caption{Transformer model with lr = 0.0003}\label{fig:trans_vs_stale1}
  \end{subfigure}
  \quad
  \begin{subfigure}[b]{0.45\textwidth}
    \centering
    \includegraphics[height=4.2cm]{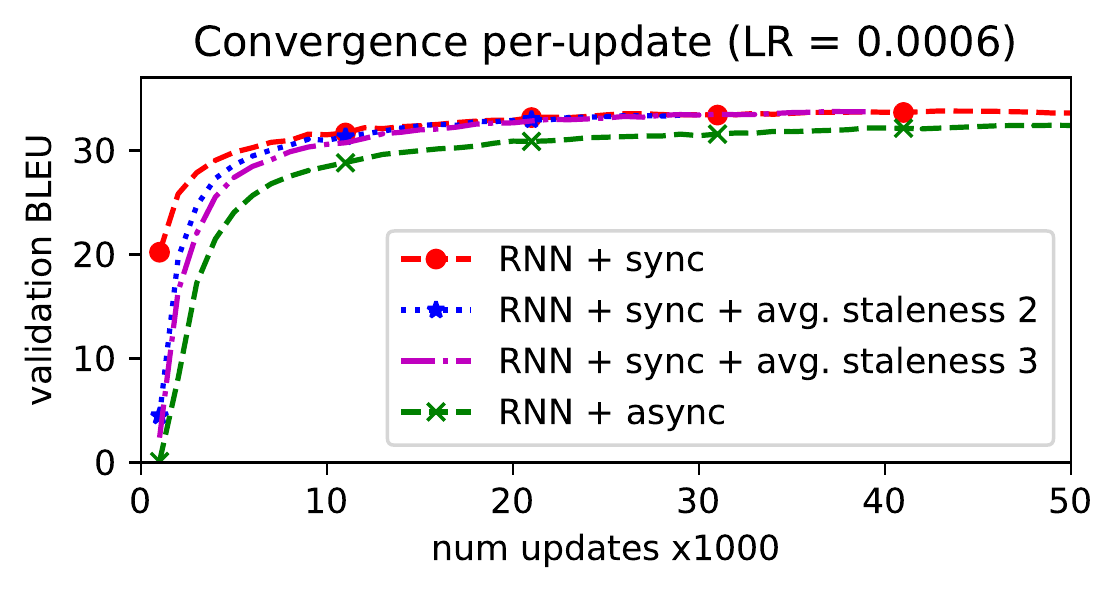}
    \caption{RNN model with lr = 0.0006}\label{fig:rnn_vs_stale1}
  \end{subfigure}
  \caption{Artificial staleness in synchronous SGD compared to synchronous and asynchronous baselines, all with our usual learning rate for each model.}
\label{fig:defaultlearn}
\end{figure*}

\begin{figure*}[t!]
\centering
  \begin{subfigure}[b]{0.45\textwidth}
    \centering
    \includegraphics[height=4.2cm]{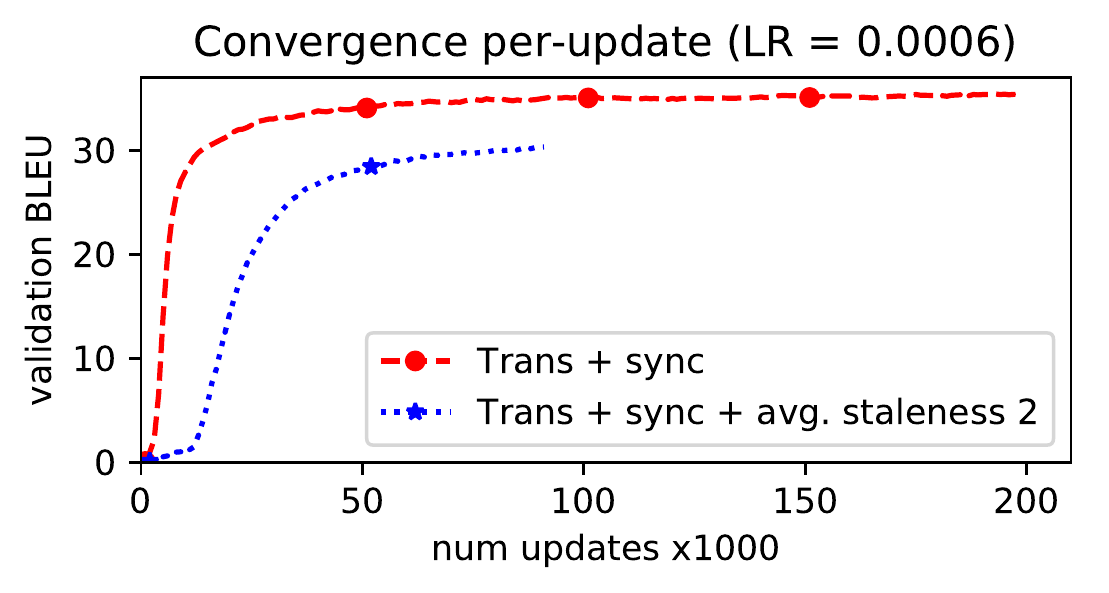}
    \caption{Transformer model with lr = 0.0006}\label{fig:trans_vs_stale2}%TODO(alham): this key really needs to have all 4 rows!!!
  \end{subfigure}
  \quad
  \begin{subfigure}[b]{0.45\textwidth}
    \centering
    \includegraphics[height=4.2cm]{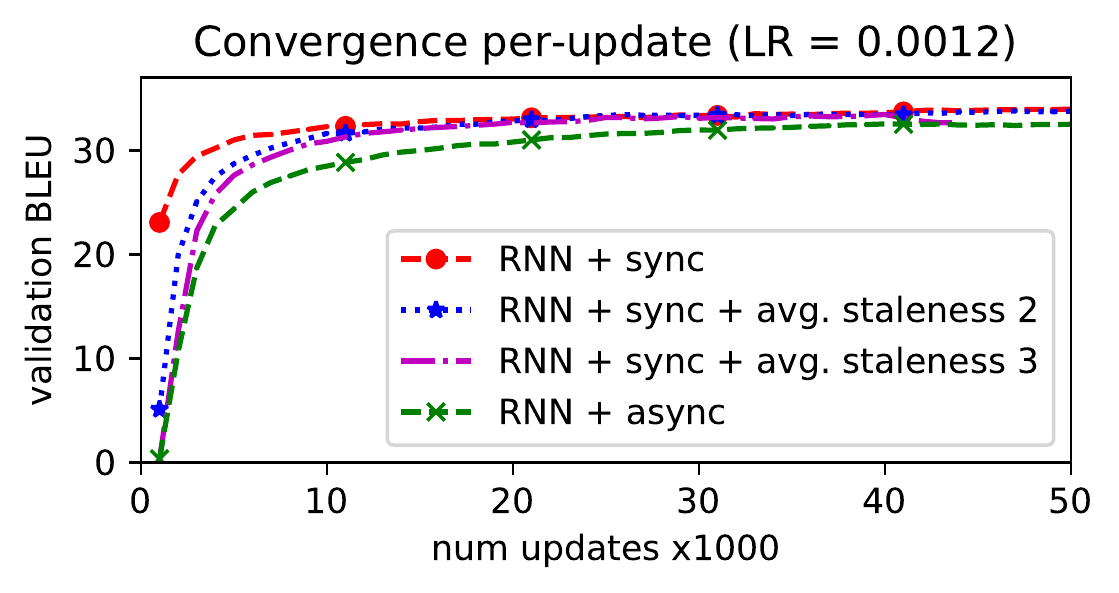}
    \caption{RNN model with lr = 0.0012}\label{fig:rnn_vs_stale2}
  \end{subfigure}
  \caption{Artificial staleness in synchronous SGD with doubled learning rates.  Transformers with learning rate 0.0006 and staleness 3 (synchronous and asynchronous) did not rise above 0.}
\label{fig:doublelearn}
\end{figure*}

A stale gradient occurs when parameters have updated while a processor was computing its gradient.  Staleness can be defined as the number of updates that occurred between the processor pulling parameters and pushing its gradient.   Under the ideal case where every processor spends equal time to process a batch, asynchronous SGD with $N$ processors produces gradients with staleness $N-1$. Empirically, we can also expect an average staleness of $N - 1$ with normally distributed computation time \cite{zhang2015staleness}.

To isolate the impact of staleness, we introduce staleness into synchronous SGD. Workers only pull the latest parameter once every $U$ updates, yielding an average staleness of $\frac{(U-1)}{2}$.  Since asynchronous SGD has average staleness $3$ with $N=4$ GPUs, we set $U = 7$ to achieve the same average staleness of $3$. Additionally, we also tried a lower average staleness of $2$ by setting $U = 5$.

In order to focus on the impact of the staleness, we set the batch size to 40 GB total RAM consumption, be they 4 GPUs with 10 GB each in synchronous SGD or emulated 40 GB batches on each GPU in asynchronous SGD.

Results are shown in Figure~\ref{fig:defaultlearn}.  Staleness 3 substantially degrades Transformer convergence and final quality (Figure~\ref{fig:trans_vs_stale1}). However, the impact of staleness 2 is relatively minor. We also continue to see that Transformers are more sensitive than RNNs to training conditions.  

\khdone{Go over results of the preceding paragraph first.  That's already a lot to swallow. Digest those results then you can move on to the learning rate stuff here. And explain more.}An alternative way to interpret staleness is the distance between the parameters with which the gradient was computed and the parameters being updated by the gradient. To see this effect, we run another set of experiments with double the learning rate, so that parameters move faster.

Results for Transformer worsen when we double the learning rate (Figure~\ref{fig:doublelearn}). With staleness 3, the model stayed at 0 BLEU for both synchronous or asynchronous SGD, consistent with our earlier result (Table~\ref{table:lr-sweep}).

% The experiment in Figure~\ref{fig:all_vs_stale} generally shows that staleness negatively affects the convergence. Similar to the batch size, staleness affects the convergence curve especially in the early stage of training. The damage is shown to be more severe under higher learning rate.

We conclude that staleness is primary, but not wholly, responsible for the poor performance of asynchronous SGD in training Transformers.  However, asynchronous SGD still underperforms synchronous SGD with artificial staleness of 3 and the same batch size (40 GB).  Our synchronous SGD training has consistent parameters across processors, whereas processors might have different parameters in asynchronous training. The staleness distribution might also play a role because staleness in asynchronous SGD follows a normal distribution~\cite{zhang2015staleness} while our synthetic staleness in synchronous SGD follows a uniform distribution.

\section{Incremental Updates in Adam}

Investigating the effect of batch size and staleness further, we analyze why it makes a difference that gradients computed from the same parameters are applied one at a time (incurring staleness) instead of summed then applied once (as in synchronous SGD).  As seen in Section~\ref{staleexperiment}, our artificial staleness was damaging to convergence even though gradients were synchronously computed with respect to the same parameters. In standard stochastic gradient descent there is no difference: gradients are multiplied by the learning rate then substracted from the parameters in either case. The Adam optimizer handles incremental updates and sums differently.

Adam is scale invariant. For example, suppose that two processors generate gradients 0.5 and 0.5 with respect to the same parameter in the first iteration. Incrementally updating with 0.5 and 0.5 is the same as updating with 1 and 1 due to scale invariance. Updating with the summed gradient, 1, will only move parameters half as far. This is the theory underlying the rule of thumb that learning rate should scale with batch size~\cite{ott2018scaling}.

In practice, gradients reported by processors are usually not the same: they are noisy estimates of the true gradient. In Table~\ref{adamslow}, we show examples where noise causes Adam to slow down. Summing gradients smooths out some of the noise. Next, we examine the formal basis for this effect.

\begin{table*}[]
\centering
\begin{tabular}{lcLLLLLLL}
\hline
\multicolumn{2}{c}{Time ($t$)}  & 0 & 1      & 2      & 3      & 4      & 5      & 6      \\ \hline
Constant        & $g_t$  & & 1      & 1      & 1      & 1      & 1      & 1      \\
                & $m_t$  & 0 & 0.1    & 0.19   & 0.271  & 0.344  & 0.41   & 0.469  \\
                & $v_t$  & 0 & 0.02   & 0.04   & 0.059  & 0.078  & 0.096  & 0.114  \\
                & $\hat{m}_t$ & 0 & 1      & 1      & 1      & 1      & 1      & 1      \\
                & $\hat{v}_t$ & 0 & 1      & 1      & 1      & 1      & 1      & 1      \\
                & $\theta$ & 0 & -0.001 & -0.002 & -0.003 & -0.004 & -0.005 & -0.006 \\
\hline
Scaled          & $g_t$  & & 0.5    & 1.5    & 0.5    & 1.5    & 0.5    & 1.5    \\
                & $m_t$  & 0 & 0.05   & 0.195  & 0.226  & 0.353  & 0.368  & 0.481  \\
                & $v_t$  & 0 & 0.005  & 0.05   & 0.054  & 0.098  & 0.101  & 0.144  \\
                & $\hat{m}_t$ & 0 & 0.5    & 1.026  & 0.832  & 1.026  & 0.898  & 1.026  \\
                & $\hat{v}_t$ & 0 & 0.25   & 1.26   & 0.917  & 1.26   & 1.05   & 1.26   \\
                & $\theta$ & 0 & -0.001 & -0.002 & -0.003 & -0.004 & -0.005 & -0.005 \\
\hline
Different sign  & $g_t$  &  & -1     & 2      & -1     & 2      & -1     & 2      \\
                & $m_t$  & 0 & -0.1   & 0.11   & -0.001 & 0.199  & 0.079  & 0.271  \\
                & $v_t$  & 0 & 0.02   & 0.1    & 0.118  & 0.195  & 0.211  & 0.287  \\
                & $\hat{m}_t$ & 0 & -1     & 0.579  & -0.004 & 0.579  & 0.193  & 0.579  \\
                & $\hat{v}_t$ & 0 & 1      & 2.515  & 2      & 2.515  & 2.2    & 2.515  \\
                & $\theta$ & 0 & 0.001  & 0.001  & 0.001  & 0.000      & 0.000      & -0.000  \\
\hline
\end{tabular}
\caption{\label{adamslow}The Adam optimizer slows down when gradients have larger variance even if they have the same average, in this case 1. When alternating between $-1$ and $2$, Adam takes 6 steps before the parameter has the correct sign.  Updates can even slow down if gradients point in the same direction but have different scales.  The learning rate is $\alpha = 0.001$.}
\end{table*}

%Adam was designed on the assumption that each gradient was taken with respect to the parameters it produced in the previous time step \cite{kingma2014adam}, which both asynchronous SGD and the proposed incremental synchronous SGD violate.  Informally, Adam cannot tell if changes in the gradient are due to noise or due to legitimate responses to parameters moving, so it slows down to avoid overshooting when gradients change.  However, in this case gradients were computed with respect to the same parameters, so changes are due entirely to noise in estimating the true gradient. Summing smooths out some of the noise.  

Formally, Adam estimates the full gradient with an exponentially decaying average $m_t$ of gradients $g_t$.
\[m_t\gets \beta_1 m_{t-1} + (1-\beta_1)g_t\]
%where $\beta_1$ is a decay hyperparameter.  It also computes a decaying average $v_t$ of second moments
\[v_t \gets \beta_2 v_{t-1} + (1-\beta_2)g_t^2\]
% where $\beta_2$ is a separate decay hyperparameter. The squaring $g_t^2$ is taken element-wise.  These estimates are biased because the decaying averages were initialized to zero.  Adam corrects for the bias to obtain unbiased estimates $\hat{m}_t$ and $\hat{v}_t$.  
\[\hat{m}_t \gets m_t/(1-\beta_1^t)\]\[\hat{v}_t \gets v_t/(1-\beta_2^t)\]
%These estimates are used to update parameters $\theta$
Parameter update:
\[\theta_t \gets \theta_{t-1} - \alpha \frac{\hat{m}_t}{\sqrt{\hat{v}_t} + \epsilon}\]
where $\alpha$ is the learning rate hyperparameter and $\epsilon$ prevents element-wise division by zero.  %We use $\beta_1 = 0.9$, $\beta_2 = 0.98$, $\epsilon = 10^{-8}$, and sweep $\alpha \in [0.0002, 0.00375]$. 
%\all{We already mentioned this in experiment section.. }

Replacing estimators in the update rule with statistics they estimate and ignoring the usually-minor $\epsilon$
\[\frac{\hat{m}_t}{\sqrt{\hat{v}_t} + \epsilon} \approx \frac{Eg_t}{\sqrt{E(g_t^2)}}\]
which expands following the variance identity
\[\frac{Eg_t}{\sqrt{E(g_t^2)}} = \frac{Eg_t}{\sqrt{Var(g_t) + (Eg_t)^2}}\]
Dividing both the numerator and denominator by $|Eg_t|$, we obtain
\[= \frac{\text{sign}(Eg_t)}{\sqrt{Var(g_t)/(Eg_t)^2 + 1}}\]
The term $Var(g_t)/(Eg_t)^2$ is statistical efficiency, the square of coefficient of variation.  In other words, Adam gives higher weight to gradients if historical samples have a lower coefficient of variation. The coefficient of variation of a sum of $N$ independent\footnote{Batch selection takes compute time into account, so technically noise is not independent.} samples decreases as $1/\sqrt{N}$. Hence sums (despite having less frequent updates) may actually cause Adam to move faster because they have smaller coefficient of variation. An example appears in Table~\ref{adamslow}: updating with 1 moves faster than individually applying -1 and 2.

With these results in mind, we focus on how best to sum gradients.

\section{Related Work}

\subsection{Gradient Summing}
Several papers wait and sum $P$ gradients from different workers as a way to reduce staleness.  In \newcite{chen2016revisiting}, gradients are accumulated from different processors, and whenever the $P$ gradients have been pushed, other processors cancel their process and restart from the beginning.  This is relatively wasteful since some computation is thrown out and $P-1$ processors still idle for synchronization. \newcite{gupta2016model} suggest that restarting is not necessary but processors still idle waiting for $P$ to finish.  Our proposed method follows \newcite{lian2015asynchronous} in which an update happens every time $P$ gradients have arrived and processors continually generate gradients without synchronization.  

Another direction to overcome stale gradient is to reduce its effect towards the model update. \newcite{mcmahan2014delay} dynamically adjust the learning rate depending on the staleness. \newcite{dutta2018slow} suggest to completely ignore stale gradient pushes.

\subsection{Increasing Staleness}
In the opposite direction, some work has added noise to gradients or increased staleness, typically to cut computational costs. \newcite{recht2011hogwild}~propose a lock-free asynchronous gradient update. Lossy gradient compression by bit quantization~\cite{seide20141, alistarh2017qsgd} or threshold based sparsification~\cite{aji2017sparse, lin2017deep} also introduce noisy gradient updates. On top of that, these techniques store unsent gradients to be added into the next gradient, increasing staleness for small gradients.

\newcite{dean2012large} mention that communication overload can be reduced by reducing gradient pushes and parameter synchronization frequency. In \newcite{mcmahan2017communication} work, each processor independently updates its own local model and periodically synchronize the parameter by averaging across other processors. \newcite{ott2018scaling} accumulates gradients locally, before sending it to the parameter server. \newcite{bogoychev2018accelerating} also locally accumulates the gradient, but also updates local parameters in between.

\section{Asynchronous Transformer Training}

%Accumulating the gradients in parameter server before running a parameter update can increase the effective batch size and reduce the staleness, which has shown to improve the training quality in asynchronous SGD. Formally, each processor send the computed gradient to the server, while the server only applies a parameter updates after receiving and summing $N$ gradient pushes. 
\subsection{Accumulated Asynchronous SGD}

Previous experiments have shown that increasing the batch size and reducing staleness improves the final quality of asynchronous training. Increasing the batch size can be achieved by accumulating gradients before updating. We experiment with variations on three ways to accumulate gradients:

\textbf{Local Accumulation:} Gradients can be accumulated locally in each processor before sending it to the parameter server~\cite{ott2018scaling, bogoychev2018accelerating}. This approach scales the effective batch size and reduces communication costs as the workers communicate less often. However, this approach does not reduce staleness as the parameter server updates immediately after receiving a gradient. We experiment with accumulating four gradients locally, resulting in 40 GB effective batch size.

\textbf{Global Accumulation:} Each processor sends the computed gradient to the parameter server normally. However, the parameter server holds the gradient and only updates the model after it receives multiple gradients~\cite{dean2012large, lian2015asynchronous}. This approach scales the effective batch size. On top of that, it decreases staleness as the parameter server updates less often. However, it does not reduce communication costs. We experiment with accumulating four gradients globally, resulting in 40 GB effective batch size and $0.75$ average staleness.

\begin{table*}[t!]
\centering
\begin{tabular}{lccccrrrrr}
\textbf{Transformer} \\ 
\hline
 Communication & \multicolumn{2}{c}{accumulation}  & batch & avg. & speed & best &  \multicolumn{3}{c}{hours to X BLEU} \\
 & local & global & size & staleness & (wps) & BLEU & 33 & 34 & 35 \\
 \hline
 % single & - & - & 10 GB & 0 & 12301 & 34.73 & 11.5 & 21.9 & 74.3 \\
 synchronous & 1 & 4 & 40 GB & 0 & 36029 & 35.66 & 5.3 & 7.6 & 15.6 \\
 asynchronous & 1 & 1 & 10 GB & 3 & 39883 & 30.72 & - & - & - \\
 asynchronous & 4 & 1 & 40 GB & 3 & 45177 & 30.98 & - & - & - \\
 asynchronous & 2 & 2 & 40 GB & 1.5 & 43115 & 35.68 & 4.9 & 6.8 & 15.4 \\
 asynchronous & 1 & 4 & 40 GB & 0.75 & 39514 & 35.84 & 4.6 & 6.7 & 11.4\\
  \hline
\end{tabular}

\begin{tabular}{lccccrrrrr}
\\
\textbf{RNN} \\ 
\hline
 Communication & \multicolumn{2}{c}{accumulation}  & batch & avg. & speed & best &  \multicolumn{3}{c}{hours to X BLEU} \\
 & local & global & size & staleness & (wps) & BLEU & 32 & 33 & 34 \\
 \hline
 % single & - & - & 10 GB & 0 & 12301 & 34.73 & 11.5 & 21.9 & 74.3 \\
 synchronous & 1 & 4 & 40 GB & 0 & 23054 & 34.30 & 3.6 & 6.2 & 18.8 \\
 asynchronous & 1 & 1 & 10 GB & 3 & 24683 & 33.76 & 2.7 & 5.1 & - \\
 asynchronous & 4 & 1 & 40 GB & 3 & 27090 & 33.83 & 4.1 & 6.1 & - \\
 asynchronous & 2 & 2 & 40 GB & 1.5 & 25578 & 34.20 & 3.2 & 5.9 & 13.7 \\
 asynchronous & 1 & 4 & 40 GB & 0.75 & 24312  & 34.48 & 3.1 & 5.4 & 14.5  \\
  \hline
\end{tabular}
\caption{Quality and convergence of asynchronous SGD with accumulated gradients on English to German dataset. Dashes indicate that model never reach the target BLEU.} \label{table:time}
\end{table*}

\textbf{Combined Accumulation:} Local and global accumulation can be combined to gain the benefits of both: reduced communication cost and reduced average staleness. In this approach, gradients are accumulated locally in each processor before being sent. The parameter server also waits and accumulates gradients before running an optimizer. We accumulate two gradients both locally and globally. This yields in 40 GB effective batch size and $1.5$ average staleness.

\begin{table*}[t!]
\centering
\begin{tabular}{lrrrrrr}
\hline
 Model & \multicolumn{2}{c}{\underline{EN $\rightarrow$ DE}} & \multicolumn{2}{c}{\underline{EN $\rightarrow$ FI}} & \multicolumn{2}{c}{\underline{EN $\rightarrow$ RU}} \\

 newstest & 2016 & 2017 & 2017 & 2018 & 2015 & 2018 \\
 \hline
 Trans. + synchronous SGD & 35.66 & 28.81 & 18.47 & 14.03 & 29.31 & 25.49\\
 Trans. + asynchronous SGD & 30.72 &  24.68 & 11.63 & 8.73 & 21.12 & 17.78\\
 Trans. + asynchronous SGD + 4x global accum. & 35.84 & 28.66 & 18.47  & 13.78 & 29.12 & 25.25 \\
 \hline
 RNN + synchronous SGD  & 34.30 & 27.43 & 16.94 & 12.75 & 26.96 & 23.11 \\
 RNN + asynchronous SGD & 33.76 & 26.84 & 14.94 & 10.96 & 26.39 & 22.48 \\
 RNN. + asynchronous SGD + 4x global accum. & 34.48 & 27.56 & 17.05 & 12.76 & 27.15 & 23.41  \\
 \hline
\end{tabular}

\caption{The effect of global accumulation on translation quality for different language pairs on development and test set, measured with BLEU score.} \label{table:end-to-end}
\end{table*}

We tested the three gradient accumulation flavors on the English-to-German task with both Transformer and RNN models. Synchronous SGD also appears as a baseline.  To compare results, we report best BLEU, raw training speed, and time needed to reach several BLEU checkpoints. Results are shown in Table~\ref{table:time}. 

Asynchronous SGD with global accumulation actually improves the final quality of the model over synchronous SGD, albeit not meaningfully.  This one change, accumulating every 4 gradients (the number of GPUs), restores quality in asynchronous methods.  It also achieves the fastest time to reach near-convergence BLEU in both Transformer and RNN.

While using local accumulation provides even faster raw speed, the model produces the worst quality among the other accumulation techniques. Asynchronous SGD with 4x local accumulation is essentially just ordinary asynchronous SGD with 4x larger batch size and 4x less update frequency. In particular, gradient staleness is still the same, therefore this does not help the convergence per-update.

Combined accumulation performs somewhat in the middle. It does not converge as fast as asynchronous SGD with full global accumulation but not as poor as asynchronous SGD with full local accumulation. Its speed is also in between, reflecting communication costs.

%The RNN model, on the other hand, is less sensitive toward stale gradients. Hence, we can accumulate some of the gradients locally for better speed without sacrificing the quality.

\subsection{Generalization Across Learning Rates}

Earlier in Table~\ref{table:lr-sweep} we show that asynchronous Transformer learning is very sensitive towards the learning rate. In this experiment, we use an asynchronous SGD with global gradient accumulation to train English-to-German on different learning rates. We compare our result with vanilla synchronous and vanilla asynchronous SGD.

\begin{table}[h!]
\centering
\begin{tabular}{lrrr}
\hline
  & \multicolumn{3}{c}{Communication}  \\ 
 & Sync. & Async. & Async \\
  Learn Rate & & & + 4x GA \\ \hline
 0.0003 & 35.66 & 30.72 & 35.84 \\ 
 0.0006 & 35.42 & 0.00 & 35.81  \\ 
 0.0012 & 33.96 & 0.00 & 33.62 \\
 0.0024 & 29.35 & 0.00 & 1.20 \\
 \hline
\end{tabular}
\caption{Performance of the asynchronous Transformer on English to German with 4x Global accumulations (GA) across different learning rates on development set measured with BLEU score.}
\label{table:lr-sweep-global}
\end{table}

Our finding empirically show that asynchronous Transformer training while globally accumulating the gradients is significantly more robust. As shown in Table~\ref{table:lr-sweep-global}, the model is now capable to learn on higher learning rate and yield comparable results compared to its synchronous variant.

\subsection{Generalization Across Languages}

To test whether our findings on English-to-German generalize, we train two more translation systems using globally accumulated gradients. Specifically, we train English to Finnish (EN $\rightarrow$ FI) and English to Russian (EN $\rightarrow$ RU) models for the WMT 2018 task \cite{wmt2018finding}. We validate our model on newstest2015 for EN $\rightarrow$ FI and newstest2017 for EN $\rightarrow$ RU. Then, we test our model on newstest2017 for EN $\rightarrow$ DE and newstest2018 for both EN $\rightarrow$ FI and EN $\rightarrow$ RU. The same network structures and hyperparameters are used as before.  

The results shown in Table \ref{table:end-to-end} empirically confirm that accumulating the gradient to obtain a larger batch size and a lower staleness in Transformer massively improves the result, compared to basic asynchronous SGD (+6 BLEU on average). The improvement is smaller in RNN experiment, but still substantial (+1 BLEU on average). We also have further confirmation that training a Transformer model with normal asynchronous SGD is impractical.

% As illustrated in Figure X, accumulating the gradients in processor reduces the communication cost as each processor communicates less often. However, accumulating gradients in the server reduces the staleness. In Transformer training, there is no benefit of faster communication as the convergence quality is bad due to the stale gradients.

\section{Conclusion}

We evaluated the behavior of Transformer and RNN models under asynchronous training. We divide our analysis based on two main different aspects in asynchronous training: batch size and stale gradient. Our experimental results show that:

\begin{itemize}

    \item In general, asynchronous training damages the final BLEU of the NMT model. However, we found that the damage with the Transformer is significantly more severe. In addition, asynchronous training also requires a smaller learning rate to perform well.

    \item With the same number of processors, asynchronous SGD has a smaller effective batch size. We empirically show that training under a larger batch size setting can slightly improves the convergence. However, the improvement is very minimal. The result in asynchronous Transformer model is subpar, even with a larger batch size.

    \item Stale gradients play a bigger role in the training performance of asynchronous Transformer. We have shown that the Transformer model's performed poorly by adding a synthetic stale gradient.

\end{itemize}

Based on these findings, we suggest applying a modification in asynchronous training by accumulating a few gradients (for example for the number of processors) in the server before applying an update. This approach increases the batch size while also reducing the average staleness. We empirically show that this approach combine the high quality training of synchronous SGD and high training speed of asynchronous SGD.

\bibliography{acl2019}
\bibliographystyle{acl_natbib.bst}

\appendix

\end{document}